# Training for ‘Unstable’ CNN Accelerator: A Case Study on FPGA

KouZi xing

xingkouzi@ict.ac.cn

## ABSTRACT

With the great advancements of convolution neural networks (CNN), CNN accelerators are increasingly developed and deployed in the major computing systems including the data centers and mobile phones. To make use of the CNN accelerators, CNN models are trained via the off-line training systems such as Caffe, Pytorch and Tensorflow on multi-core CPUs and GPUs first and then compiled to the target accelerators. Although the two-step process seems to be natural and has been widely applied, it assumes that the accelerators’ behavior can be fully modeled on CPUs and GPUs. This does not hold true and the behavior of the CNN accelerators is un-deterministic when the circuit works at ‘unstable’ mode when it is overclocked or is affected by the environment like fault-prone aerospace. The exact behaviors of the accelerators are determined by both the chip fabrication and the working environment or status. In this case, applying the conventional off-line training result to the accelerators directly may lead to considerable accuracy loss.

To address this problem, we propose to train for the ‘unstable’ CNN accelerator and have the ‘un-determined behavior’ learned together with the data in the same framework. Basically, it starts from the off-line trained model and then integrates the uncertain circuit behaviors into the CNN models through additional accelerator-specific training. The fine-tuned training makes the CNN models less sensitive to the circuit uncertainty. We apply the design method to both an overclocked CNN accelerator and a faulty accelerator. According to our experiments on a subset of ImageNet, the accelerator-specific training can improve the top 5 accuracy up to 3.4% and 2.4% on average when the CNN accelerator is at extreme overclocking. When the accelerator is exposed to a faulty environment, the top 5 accuracy improves up to 6.8% and 4.28% on average under the most severe fault injection.

## 1 Introduction

Convolutional neural network (CNN) has received huge attentions in various applications such as video surveillance, image searching, speech recognition, and robot vision in the past few years. As CNN is usually computing intensive and critical to the system, a large number of customized CNN accelerators ([2][9][10][12][13][18]) have been developed for the sake of both higher energy-efficiency and lower processing latency on either ASICs or FPGAs. It is expected that CNN processing will increasingly be offloaded to the accelerators from the general purposed processors.

In order to make use of the CNN accelerator in the applications, a neural network is usually trained using frameworks like Caffe, Pytorch and Tensorflow on general computing systems first to get the corresponding CNN model. Then the CNN model is applied to the CNN accelerator via either a compiler or reconfiguration. The two steps are relatively independent while the implicit assumption is that the training system can produce equivalent computing result to the CNN accelerator. It holds true for conventional acceleration systems and the off-line training ([7][8][14][17]) is sufficient for most computing systems.

However, it is not the case for CNN accelerators with approximate circuits or un-deterministic behaviors. For instance, when the CNN accelerator is overclocked ([3][15]), circuit behavior becomes un-deterministic but still works with even higher performance. Under such a circumstance, the off-line training can no longer model the computing on the accelerator precisely. If the training process ignores the accelerators’ dynamic behavior, the prediction accuracy of the resulting model may degrade dramatically. Another typical occasion is the soft error like single event upset (SEU) in the accelerator. The soft error make it difficult for the general purposed processors to model the exact behavior of the CNN accelerator. To some extent, it is similar to overclocking effect from the perspective of training.

To address the above problems, we propose to train with the ‘unstable’ accelerator to tolerate the accelerators’ un-deterministic behavior. Then we revisit the conventional training flow, define the interface to integrate the hardware accelerator into the Caffe framework and performs the necessary modification to the general CNN accelerator design. Finally, we take the overclocked and fault-prone CNN accelerator as examples and demonstrate the potential of this training method. Basically, it helps to hide the low-level circuit influence to higher level applications. For the overclocking case, the application gets improved performance without much consideration of the side effects. For the accelerator exposed to soft errors, the system can proceed using the CNN model without dealing with the soft error.

The contribution of this work is summarized as the following aspects.

- We proposed to train for the ‘unstable’ CNN accelerator such that the resulting model can learn the underlying “un-deterministic circuit behavior” together with the application data. With this method, the resulting system can tolerate the CNN accelerators’ un-deterministic behavior without hardware modification.

- We build an open-sourced end-to-end training framework based on Caffe to train for the 'unstable' accelerator. In addition, we present the necessary modification of the general CNN accelerators to make use of the training framework.
- We take overclocking and fault-prone CNN accelerator as two examples and demonstrate the usefulness of the proposed system.

The paper is organized as follows. Section II analyzes the influence of the CNN accelerator's 'un-deterministic' behaviors when using the conventional training. Section III presents the proposed training of both data and the accelerator in the same framework. Meanwhile, the necessary modification of the general CNN accelerator for taking advantage of the training is detailed. Afterwards, we commit two case study using the proposed training system and demonstrate the usefulness of the system. Section IV briefs the related work. Section V draws the conclusion.

## 2. Motivation

As the major training systems typically adopt the off-line training method on CPUs and GPUs, 'un-deterministic' behaviors of the accelerators will not be considered by default. To evaluate the influence, we take an overclocked CNN accelerator and a soft-error affected CNN accelerators as examples and investigate the influence of the 'unstable' circuit on the prediction accuracy in this section.

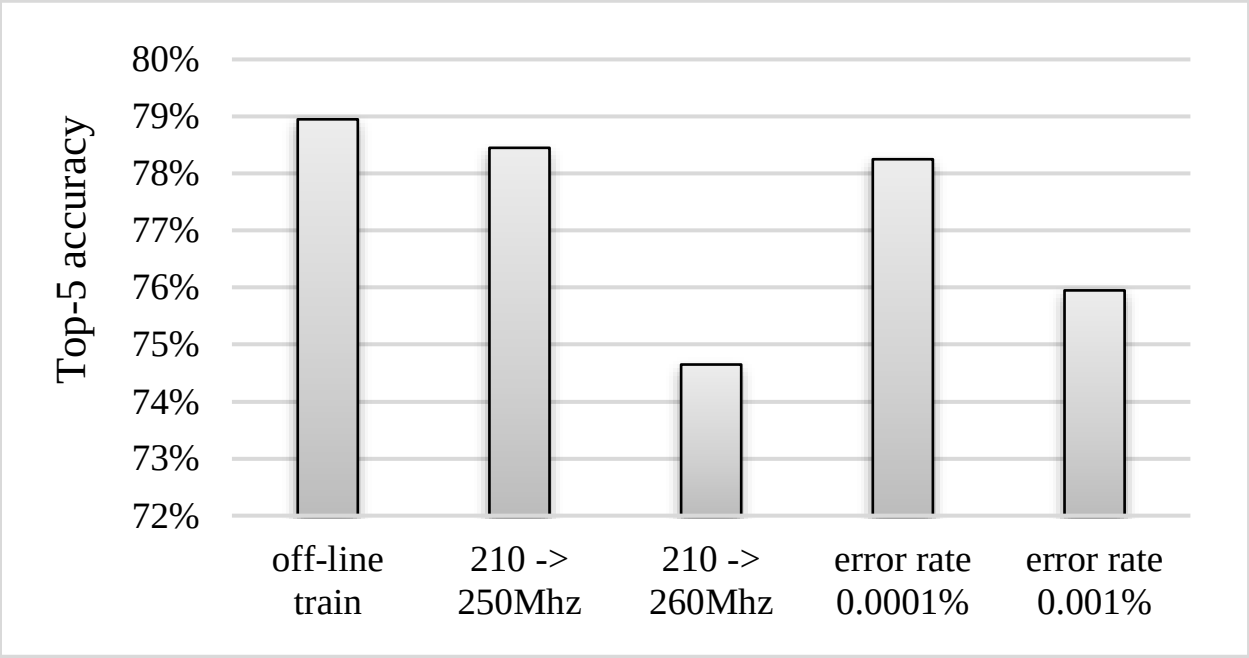

**Figure 1: Influence of The CNN Accelerator's Overclocking and Soft error on AlexNet Model Top-5 Accuracy**

As shown in Figure 1, we adopt PipeCNN[2] , an open sourced CNN accelerator, as the baseline accelerator and implement it on Xilinx KCU1500 boards. The accelerator runs at most 210 MHz safely for AlexNet. On a subset of ImageNet, we train AlexNet off-line and then apply it to the accelerator. The resulting top-5 accuracy is 78.95%. Then the clock frequency is boosted to 250Mhz and 260MHz respectively, we apply the original model on the overclocked accelerator. The performance gets improved proportionally, but the accuracy drops 0.5% and 4.3% respectively.

Soft error has become an un-ignorable problem with the shrinking semiconductor feature size and we further analyze its influence on the CNN accelerator. Currently, we use a uniform distribution model to inject the SEU errors to the multiplication-accumulation operators (MAC) of the accelerator. It causes one-bit flip on random bits of the MAC results. When the error injection rate is set to be 0.001% per MAC, applying the off-line trained model to the CNN accelerator leads to around 0.7% accuracy loss of the top 5 prediction. When error injection rate is set to be 0.001% per MAC, the prediction accuracy drops around 3%.

To gain insight on the precision drop, we also check the output of the AlexNet's last layer. We compare the data and find that the data deviates from expected result slightly due to the 'unstable' accelerator. But the accuracy loss of the model deployed on the 'unstable' CNN accelerator cannot be ignored according to the experiments. And it is highly demanded to explore the training system and take the accelerators' behavior into consideration during training for higher prediction accuracy.

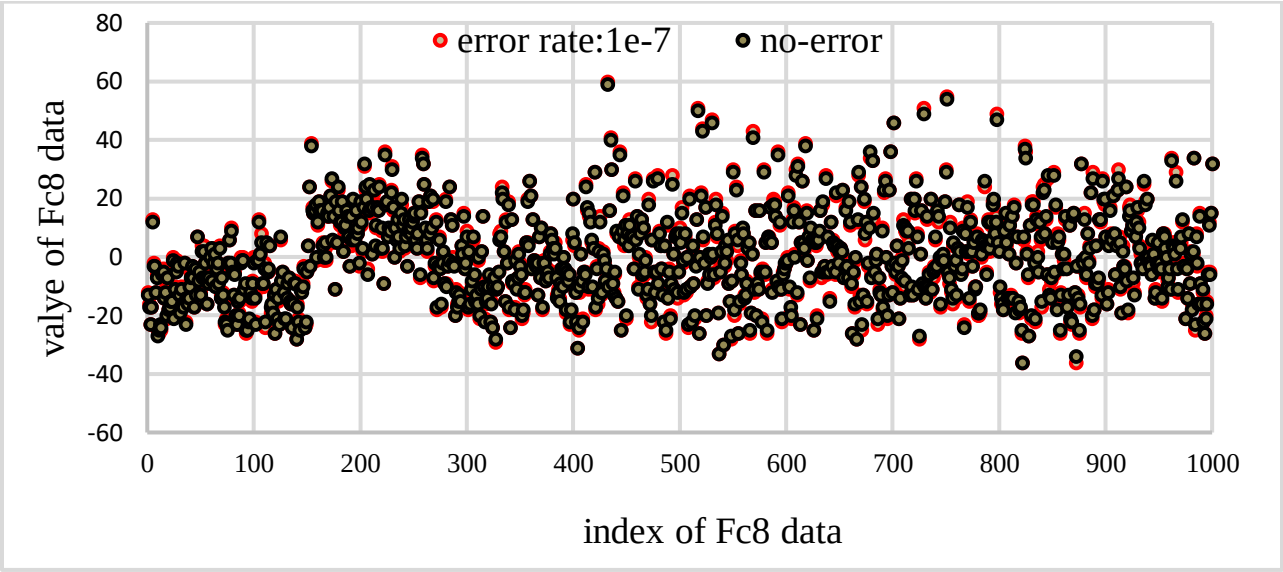

(a). soft error, 1e-6

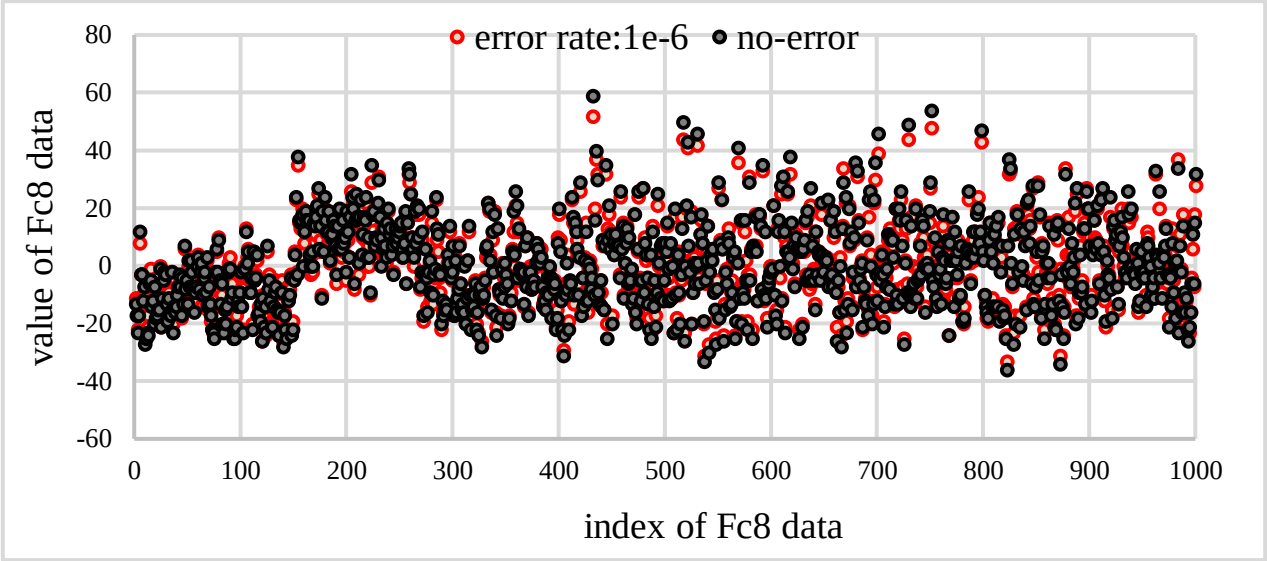

(b). soft error, 1e-5

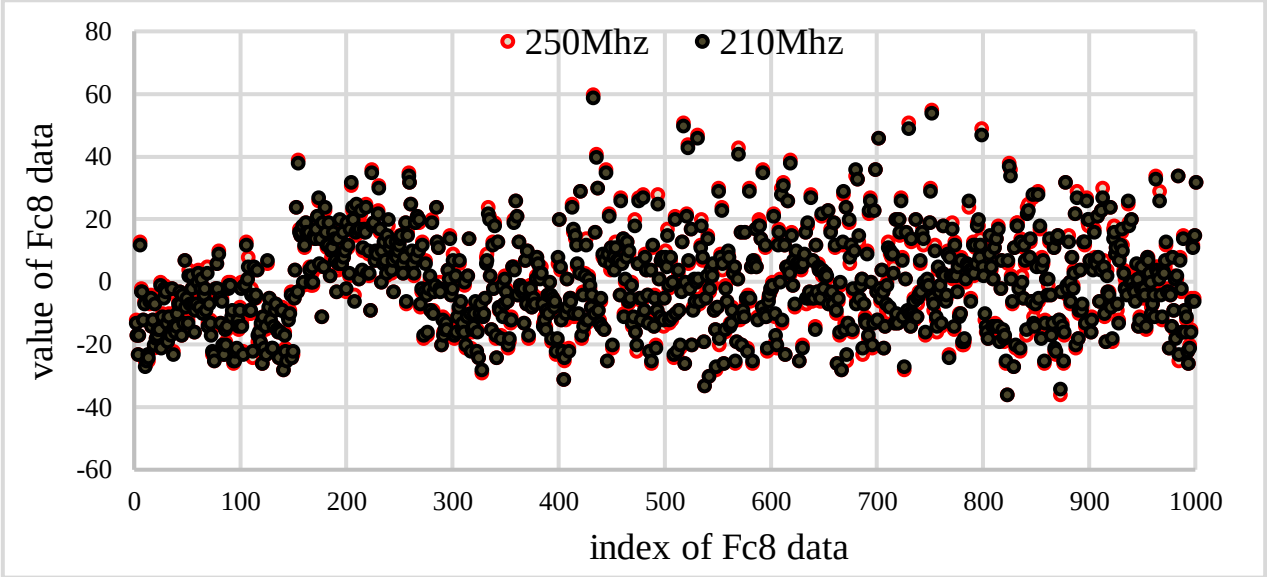

(c). overclocked, 250Mhz

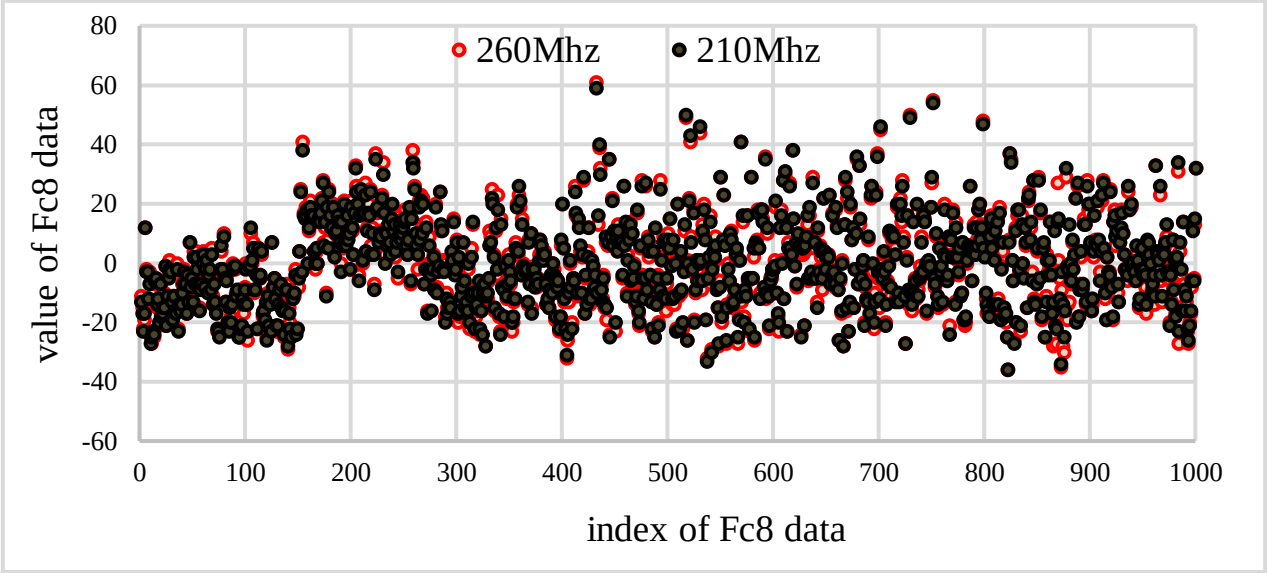

(d). overclocked, 260Mhz

**Figure 2. The Distribution of Last Layer under Overclocked and Soft error**

## 3. Training for 'Unstable' CNN accelerator

In contrast to the conventional off-line training on CPUs and GPUs, we propose to take these accelerators' dynamic behaviors into consideration during training to tolerate the 'un-deterministic' behavior of the 'unstable' accelerators. The basic idea is to embed the CNN accelerator into the conventional training framework so that the accelerator is referenced during training. In this work, we choose Caffe as the baseline training framework because it is more natural to integrate the C/C++ based high level synthesis CNN accelerator. Based on Caffe, we further detail the required general interface to make use of the hardware accelerator in training, and introduce the necessary modifications to the CNN accelerator structure.

## 3.1 Proposed Training Framework

Figure 3 illustrates the proposed training framework. It begins with the off-line training result which can greatly shorten the overall training time. While most of the CNN accelerator adopts fixed point operations, the pre-trained model is therefore expected to be fixed point model. With the pre-trained model, we mainly try to have the trained model to further adapt to the 'un-deterministic' behaviors which are difficult to model on CPUs and GPUs.

To that end, we have the forward propagation performed on the accelerator directly while the backward propagation remains on CPUs or GPUs. Forward propagation on the accelerator is fixed point, which is beneficial to both the resource consumption and memory bandwidth overhead, but backward propagation on CPUs or GPUs remains floating point to ensure the small changes in the parameters get accumulated[8]. As a result, we still need additional converting between the fixed point and floating during the training in each iteration when the weight is adjusted. When the final accuracy loss reaches the threshold, it means that the model can tolerate the accelerator's 'un-deterministic' behaviors. And the CNN model can be safely deployed on the 'unstable' accelerator.

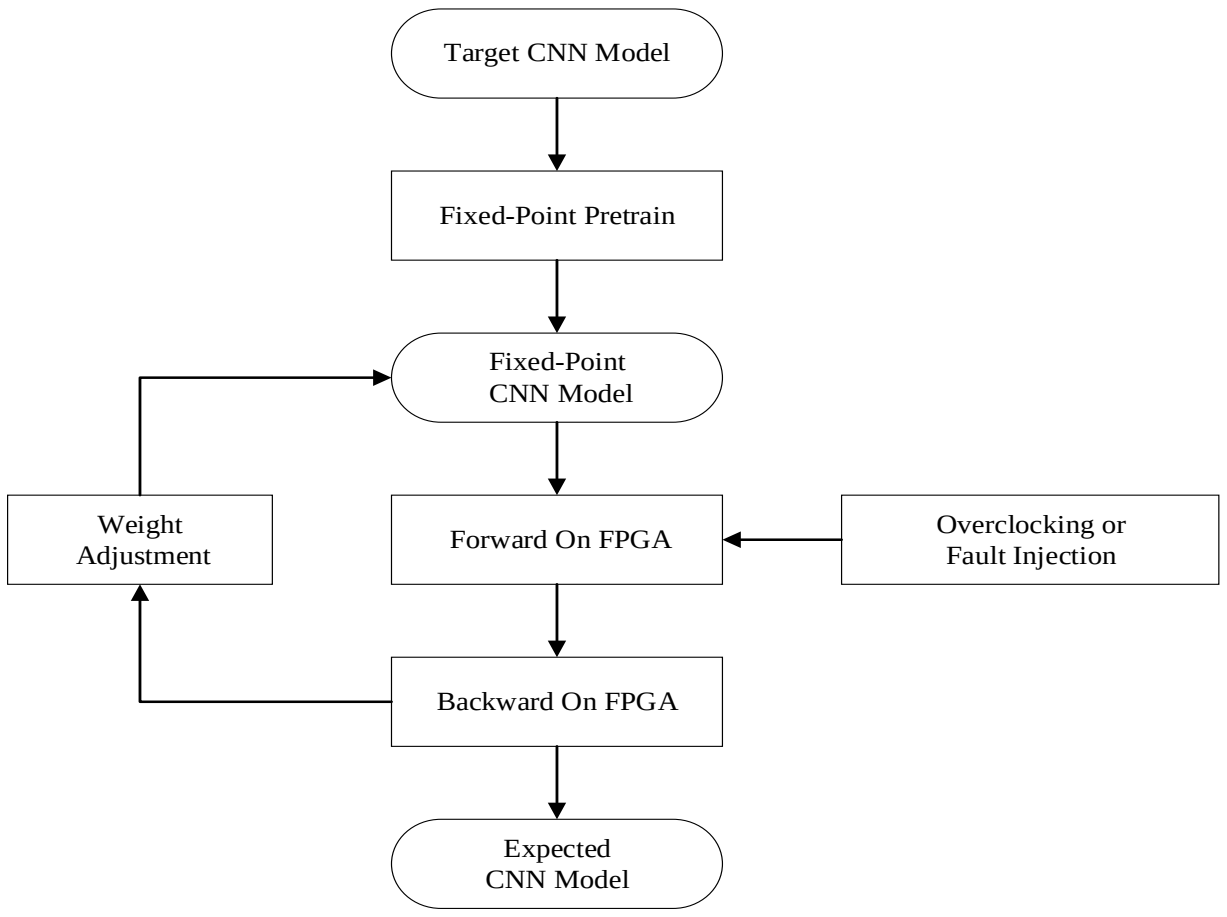

**Figure 3. Proposed Training Framework**

Figure 4 depicts the implementation of the training framework on a hybrid CPU-FPGA architecture. In this work, we use Xilinx KCU1500 as the FPGA board and put it on a standard desktop computer. CPU is the controller and it reconfigures the accelerator for a specific CNN structure. In each training iteration, CPU launches the CNN accelerator to perform the forward propagation from bottom layer to top layer. CPU does the backward propagation from top layer to bottom layer. Weights and the image data are initially stored in host memory. It will be transferred to FPGA off-chip memory for forward propagation through PCI-E. Similarly, the output data will be transferred from FPGA off-chip memory back to host memory after forward propagation. Because of the OpenCL based API wrapper in SDAccel, the CNN accelerator's interface can be easily exposed to Caffe for referring to the forward propagation result.

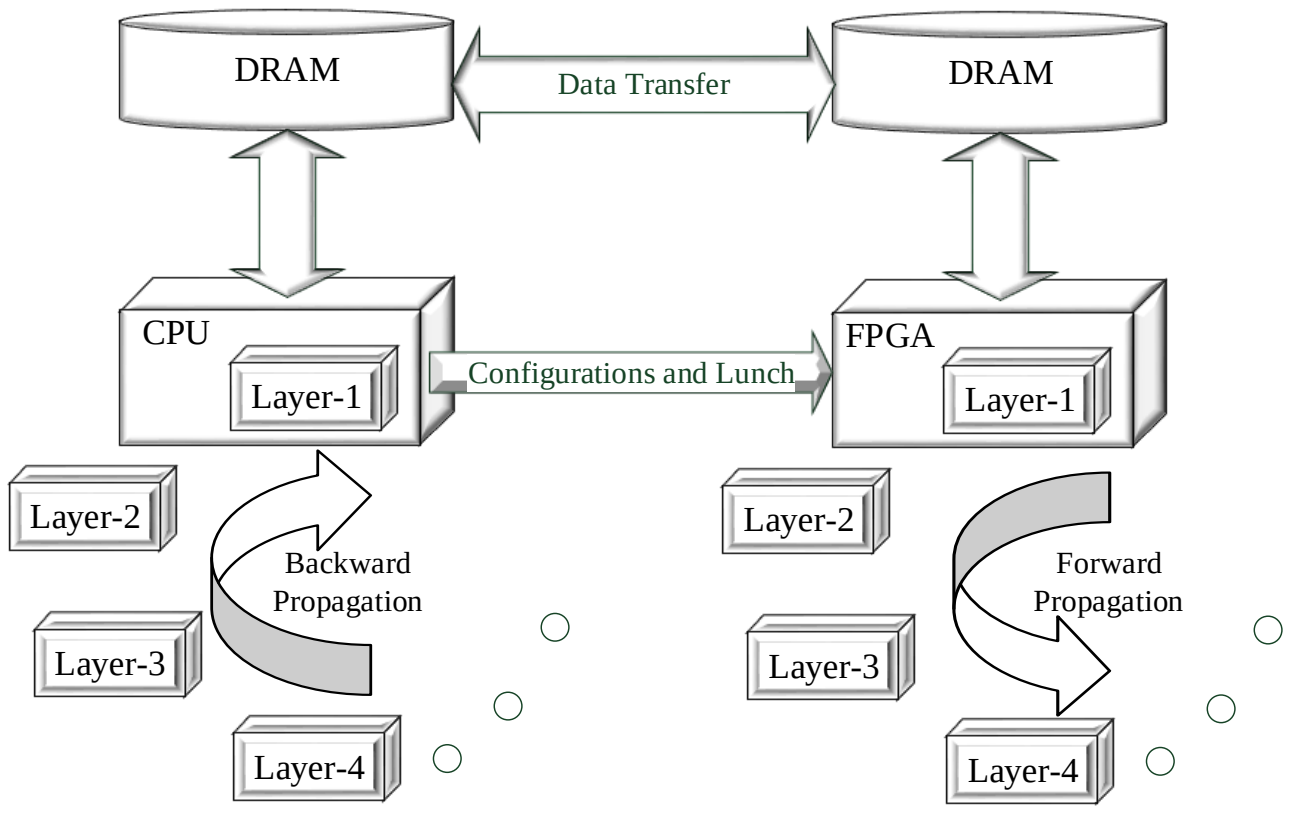

**Figure 4: Training on Hybrid CPU-FPGA Architecture**

## 3.2 High Level Accelerator Interface to Caffe

With the growing popularity of deep learning, massive different CNN accelerators have been developed over the years. In order to fit various CNN accelerators within the same training framework, we define a set of high-level interface functions as listed in Table 1. There are 7 functions included. Function 1 is used to launch the CNN accelerator from host. Function 2 and 3 are used to transfer data between the host memory and the device memory during the training. As most of the accelerators are fixed point and used for forward while back propagation is floating point, Function 4 and 5 are required for training when forward and backward propagation are iteratively committed. Function 1 to 5 are required for all the accelerators. Function 6 and 7 are only needed for accelerators that perform on reorganized data([2][12]). With the interface functions, general CNN accelerators can be trained to tolerate 'un-deterministic' circuit behaviors using the proposed framework.

CNN accelerators can either be implemented using high-level synthesis tools (HLS) or hardware description languages (HDL). With Xilinx SDAccel, we can wrap the both types of accelerators with OpenCL API. With the OpenCL API, Caffe can refer to the accelerators during training conveniently.

**Table 1. High-level interface to Caffe**

| ID | Function Name | Description |
|---|---|---|
| *1* | *launchAccelerator()* | It configures the CNN accelerator and launches it from host CPU. |
| *2* | *dataToFPGA( weight, input, wgtDevAddr, inDevAddr)* | It transfers both the input data and weight to the FPGA device memory. |
| *3* | *dataFromFPGA( outputDevAddr,* | It transfers all the intermediate output of the CNN layers from |
| | *output)* | FPGA device memory to host memory. |
| 4 | *convertIntToFloat(int iData, float fData)* | It converts the fixed-point output to float for back propagation processing. |
| 5 | *convertFloatToInt(float fData, int iData)* | It converts the floating-point input and weight data to fixed point or integer for forward processing on the accelerator. |
| 6 | *dataLayoutReorder(data, reorderedData)* | It reorders the data layout for more efficient accelerator execution before sending to FPGA device memory. |
| 7 | *dataLayoutRecover(reorderedData, data)* | It reorders the output data back to the default format for Caffe back propagation. |

### 3.3 Modification to the general CNN accelerators

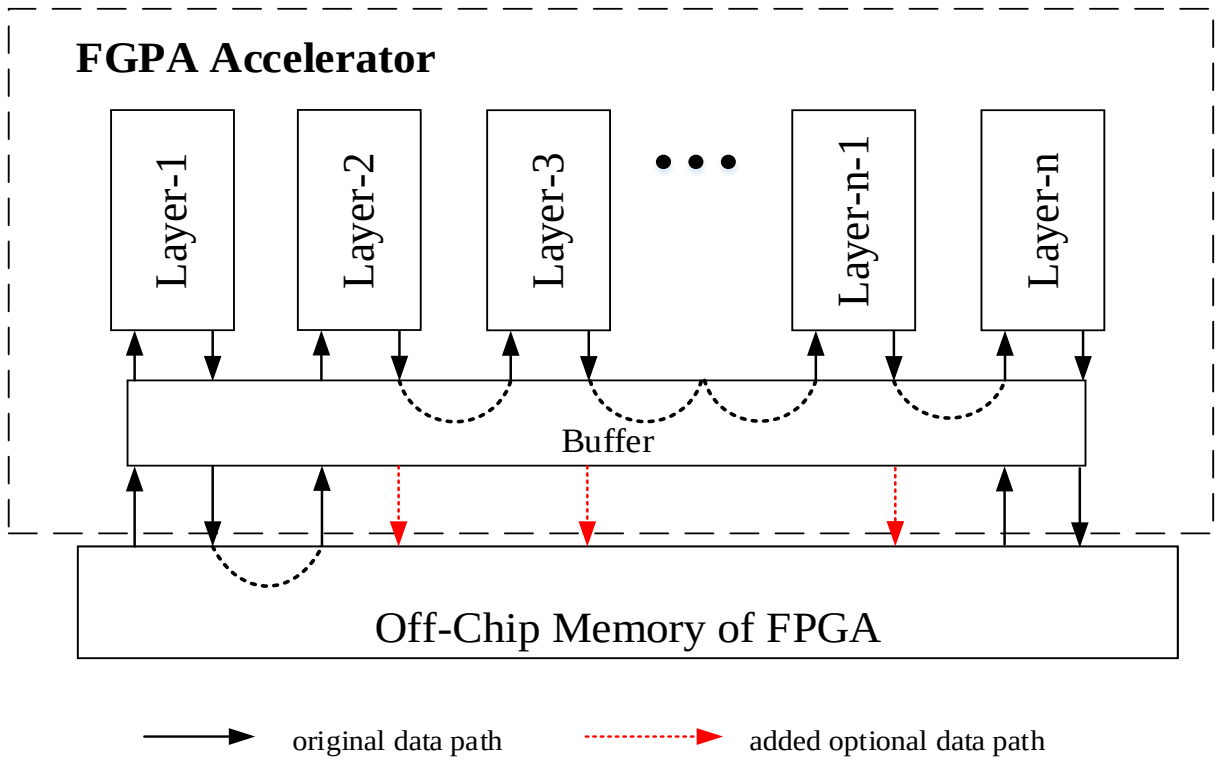

**Figure 5: Modification of the CNN accelerator data path. It essentially ensures each CNN layer to have an optional data path to off-chip memory so that it can be used for training as necessary.**

On top of the interface, the CNN accelerator also needs minor adjustment for the training. The training process requires the feature maps of each CNN layer for backward propagation, while the accelerators are typically optimized for inference and some of the layers' output are fully buffered in on-chip memory for less memory access overhead. In this case, the accelerator should make intermediate output write back optional as shown in Figure 5. When the accelerator is used in training, the output will be transferred to memory. When it is used for inference, it can also turn off the write back data path for better performance. It is trivial to modify the CNN accelerators and the hardware overhead is negligible.

## 4. Case Study and Experiments

In this section, we mainly explore deploying the CNN models on the 'unstable' accelerators using the proposed training framework. In particular, we take an overclocked CNN accelerator and a soft error attacked CNN accelerator as typical 'unstable' accelerator examples.

Four convolution neural networks including LeNet, AlexNet, VGG-16 and VGG-19 are used. They are implemented on Xilinx KCU1500 based on 8bit fixed-point PipeCNN using SDAccel 2017.1. The FPGA cards is attached to a desktop computer configured with Intel(R) Core(TM) i7-2600 CPU (4core, 3.40GHz) via PCI-e 2.0. The communication bandwidth is 8GB/s.

### 4.1 Overclocked CNN accelerator

Clock frequency is almost proportional to the computation capability of the CNN accelerator when its architecture is determined. While timing analysis tools typically recommend a conservative clock frequency in order to avoid the possibility of timing violations, FPGA designs can be safely overclocked by a significant ratio with respect to the maximum operating frequency estimated by the FPGA's tool flow. This gives the advantage of increasing the implementation throughput without any design-level modifications. Beyond this safe overclocking margin, some critical paths in the design starts to fail and the output error rate increases exponentially with respect to the increase in the clock frequency. To tolerate the computing error and gain the performance benefit, we thus opt to use the proposed training framework.

With PipeCNN, we implemented four CNN including LeNet, AlexNet, VGG-16 and VGG-19 on KCU1500. As PipeCNN provides customized implementation for different CNN, the baseline frequency of the implementations is different. The frequency of the four implementations is 210Mhz, 210MHz, 190MHz and 190MHz respectively. Then we boost the clock frequency gradually and train for the 'unstable' CNN accelerator implementations on ImageNet data set.

The prediction accuracy of the resulting implementations is presented in Figure 6. In general, overclocking can enhance the clock frequency by 19% to 26%. While applying the off-line trained model to the accelerator with overclocking, the top-5 accuracy degrades by up to 4.3%. When the proposed training framework is used, the resulting retrained model can be much better especially near the overclocking limit.

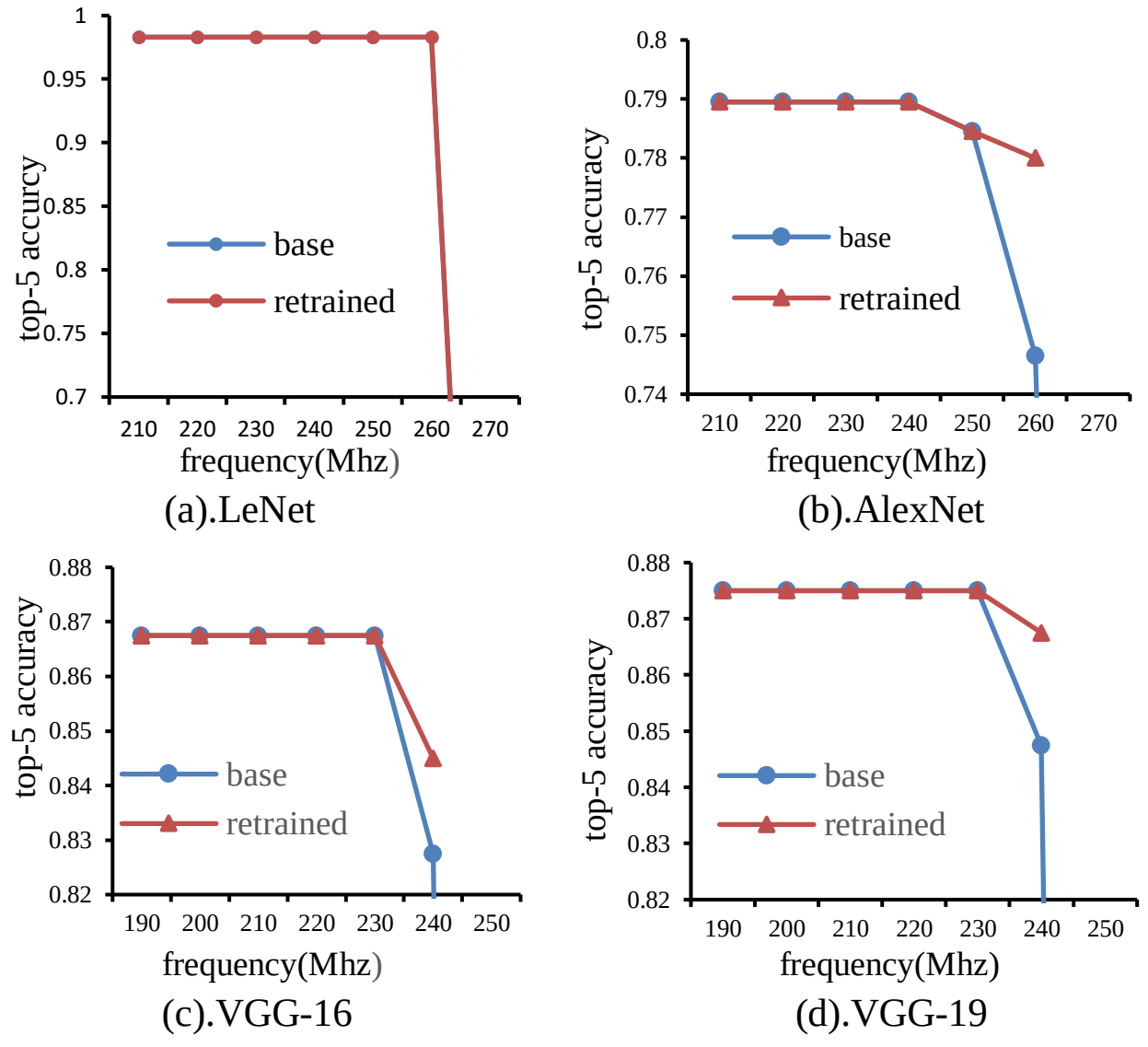

**Figure6: The top-5 Accuracy of Four CNN models on accelerators with different overclocking frequency**

For AlexNet, VGG-16 and VGG-19, the top-5 accuracy of the retained models is improved by 3.4%, 1.8%, and 2% respectively at the extreme overclocking frequency. For LeNet which is a rather small yet reliable network compared to the other three, the implementation remains unaffected even when the clock is boosted to 260 MHz from 210 MHz. When the clock goes up to 270MHz, the timing error can no longer be tolerated by the hardware system, the prediction accuracy drops to 10% which is essentially meaningless. In this case, the base model and the retrained model is pretty much the same.

To ensure the stability of the overclocking experiment, we also keep measuring the accuracy of the accelerators with extreme overclocking. With repeatedly running the test for up to 40 hours, the measured accuracy varies slightly as present in Figure 7. Despite the fact that the errors caused by the overclocking can be hardly modeled precisely at runtime, the inherent error patterns may still partly be captured by the CNN model with the proposed training. This explains the higher prediction accuracy with the retrained model. According to the above experiments, we can conclude that the proposed accelerator aware training can produce more resilient CNN model tolerating errors caused by intensive overclocking.

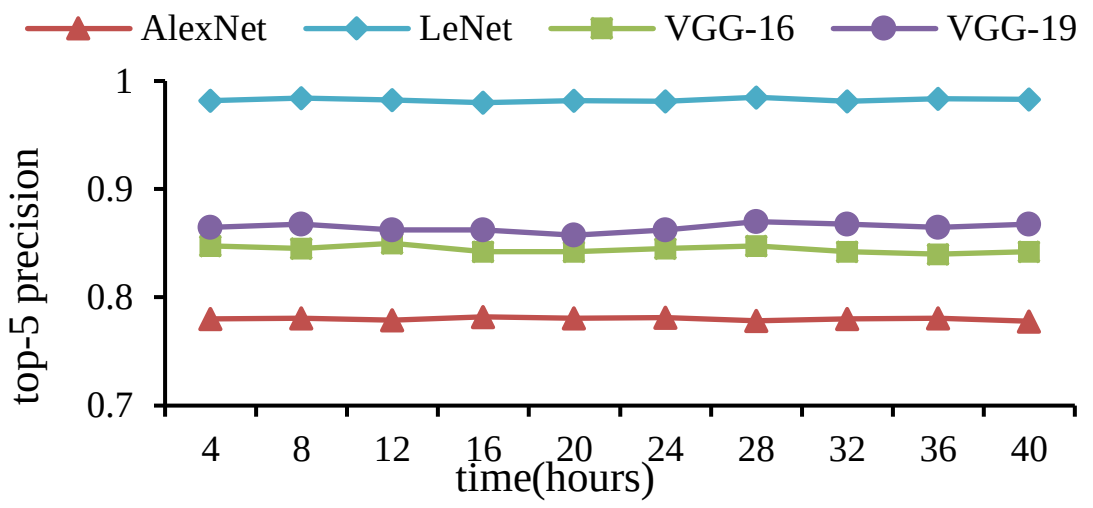

**Figure 7: The Stability of Retrained Model**

Finally, we also present the training time on the hybrid CPU-FPGA architecture. It can be seen that the training is much slower than the fixed-point training on CPU. This is mainly caused by the frequently data transferring between device memory and host memory in the proposed training, while this will not affect the inference time. In addition, we can also find that the training on larger network takes longer time and higher clock frequency is also beneficial to the training time as expected.

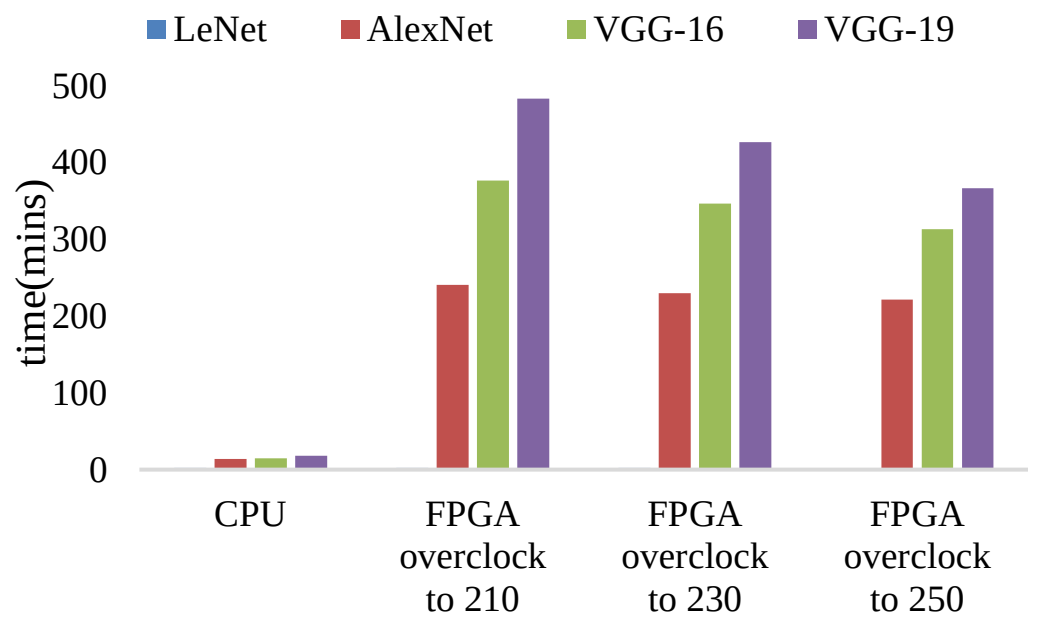

**Figure 8: Training time**

## 4.2 CNN accelerator with soft errors

With the shrinking semiconductor feature size and increasing FPGA capacity, FPGA design gets error-prone to the transient faults (often known as soft errors). They can affect the behaviors of the FPGA design dramatically. Many researchers [20][21][22][23][24] have proposed diverse approaches to address this problem. While CNN accelerators on FPGA can be different from general hardware design because the CNN model deployed can be further trained and tolerate the soft errors as proposed in prior section[1].

To explore the influence of soft errors on CNN accelerator, we need to inject soft errors to the system first. A number of fault injection techniques have been proposed in prior literature. In this work, we adopt a simple software simulation-based method to inject random errors. Although the error may be caused by on-chip memory or other SRAM cells, we have a random bit of the computing result flipped at a specific rate. The simple yet representative model will not increase the training time too much.

We also take LeNet, AlexNet, VGG-16, and VGG-19 to evaluate the influence of soft errors on prediction accuracy， the top-5 accuracy of the resulting implementations is presented in Figure 9. When we gradually increase the error rate from 1E-7 to 1E-5, the prediction accuracy degrades accordingly when applying the off-line trained model directly on the faulty accelerator. When the error rate goes up to 1E-4.5, the accuracy in the worst case drops by around 13.5%. Similar to overclocking, LeNet can tolerate more errors than the other three networks. The accuracy remains unchanged until the error rate reaches 1E-3. When the error injection rate is low, the CNN model is able to cover almost all the negative influence on the prediction accuracy.

When the error injection rate goes higher, the proposed retraining becomes critical. According to the experiments, the accuracy of the four retrained models improve by 6.8%, 3.1%, 4.25%, and 3% respectively compared to that of the base model when the accelerators are exposed to the highest error injection. In summary, the experiments demonstrate that we can have the CNN model to learn both the characteristics of the data and the underlying ‘un-deterministic’ behaviors of the accelerator together using the proposed training framework. The resulting CNN model can improve the accuracy without any modification on the accelerator when there is high error injection rate.

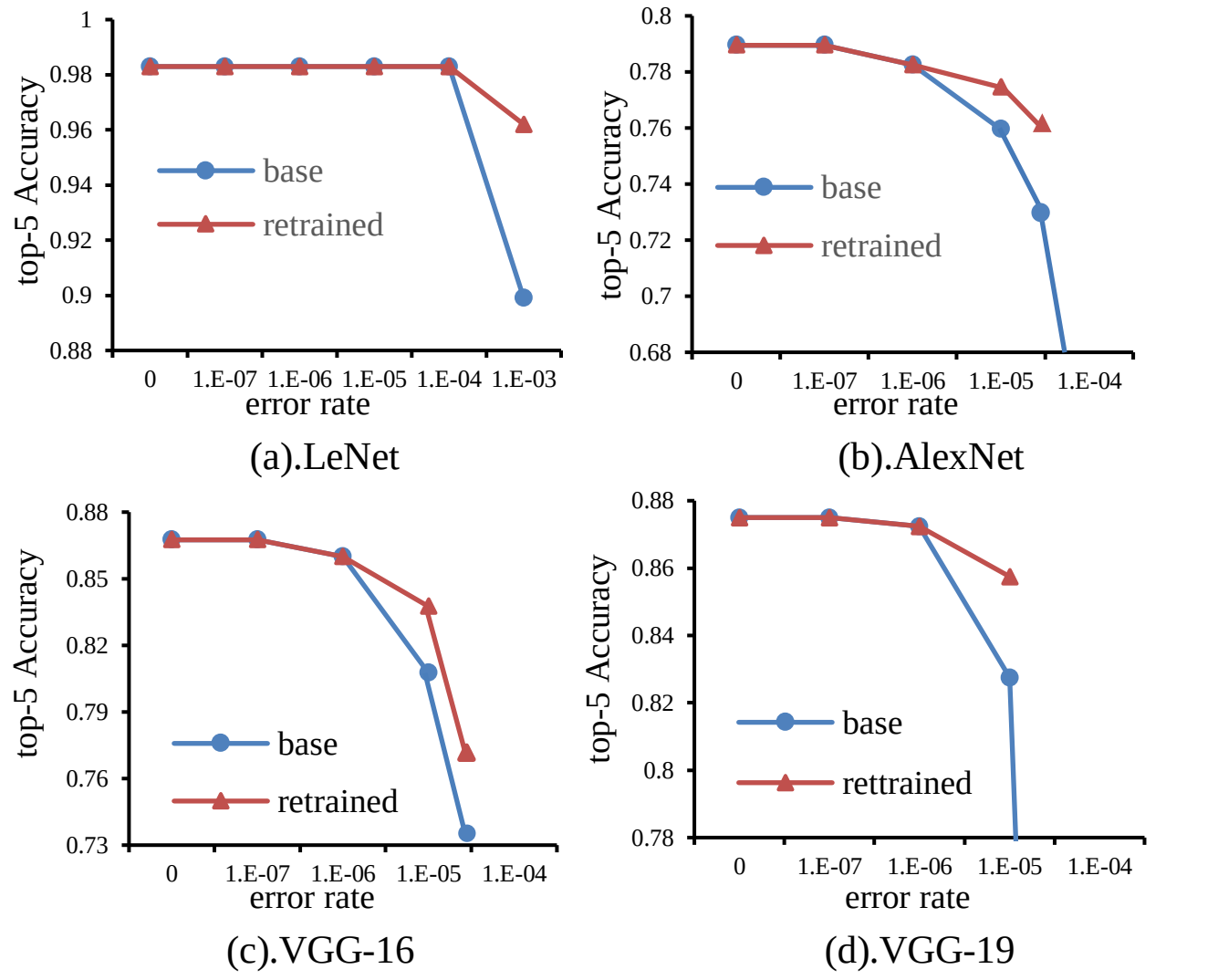

**Figure 9: The top-5 Precision of Four CNN models on accelerators with different error rate**

## 5. Related Work

**CNN accelerator:** There have been notable efforts made to create hardware accelerators of machine learning algorithms for the sake of higher performance and energy-efficiency [25] in the past few years. Among the accelerators, the regular 2D array architecture has become a mainstream solution because of the relatively higher PE and bandwidth utility. Runtime reconfigurable PE arrays are applied to provide customized solutions for efficient CNN inference on FPGAs ([6][12]). In [27], an array of processing elements (PEs) with novel architecture was developed. With intensive data reuse, it reduces the external memory bandwidth requirements dramatically and outperforms the systolic-like structure proposed in [6]. Compared to the compact hardware design in ([6][27]), Wei X et al. in[29] implemented a high-throughput CNN design and did comprehensive design space exploration on top of accurate models to determine the optimal design configuration.

**Training of accelerators:** Training approaches for CNN accelerator can be classified into three categories: (1) convert a pre-trained floating point CNN model into a fixed point model without training, (2) train a CNN model with fixed point constraint, and (3) FPGA-implemented forward & backward propagation training tools. For first category, [19] applied codebook based on scalar and vector quantization methods in order to reduce the model size. [25] analyzed the quantization sensitivity of the network for each layer and then manually decide the quantization bit-widths. [17] find direct quantization for fixed-point network design does not yield good results and optimized the fixed-point design by employing back propagation based retraining. [8] adapted a higher precision for the parameters during the updates than during the forward and backward propagations for accumulating small changes in the parameters. [17] used only binary weights to train deep neural networks.

However, these approaches of the former two categories are not suitable for 'unstable' circuit. For the third category, FCNN[5] reconfigured a streaming data path at runtime to cover the training cycle for the various layers in a CNN. Caffeine[6] provides tunable parameters, including the number and size of input/output feature maps, shape and strides of weight kernels, pooling size and stride, ReLU kernels, and the total number of CNN/DNN layers. Caffeinated FPGA[4] implemented FPGA kernels for forward and backward for Caffe and these kernels target the Xilinx SDAccel OpenCL environment for training and inference with CNNs. However, these approaches did not consider the unstable hardware behavior into their framework or either gave a way to train CNN under the un-deterministic situation.

**Unstable Hardware Behavior:** Overclocking, soft Error, circuit defect induced by process variation etc. result in the un-determined behavior of the circuits. Overclocking, a technique to gain the additional performance from a given component by increasing its operating speed, may cause timing error. [3] gave the strands of research of arithmetic precision determination and overclocking. Razor[15] projected scaled the supply voltage and clock frequency beyond the most conservative value. Soft errors are unintended transitions of logic state in a circuit typically caused external source of ionizing radiations. The shrinking transistor sizes increased the soft-errors. [20] proposed An Automated SEU Fault-Injection Method and Tool for HDL-Based Design. [30] inject single-bit flips into the register-transfer level descriptions of floating-point ALUs.

## 6. Conclusion

In this paper, we propose to take the CNN accelerator's 'un-deterministic' behaviors into consideration at training and have the CNN model to learn the accelerator's behaviors. To that end, we further build an open-sourced training system based on Caffe on a hybrid CPU-FPGA architecture. Then use the training system to deal with an overclocked CNN accelerator and an accelerator with soft errors. According to our experiments, the proposed training can improve the prediction accuracy of four CNN models up to 3.4% when the CNN accelerator is overclocked on the extreme situation. This method is also beneficial to the CNN accelerators with soft errors. In the case with most soft errors, it improves the prediction accuracy up to 6.8% and by 4.28% on average. The disadvantage is the much longer training time due to the frequent data transfer between host memory and device memory. This problem can be resolved when porting the system to closely coupled CPU-FPGA architectures with shared memory.